\documentclass[a4paper]{article}

\usepackage{url}
\usepackage{tipa}
\usepackage{INTERSPEECH2021}
\title{Is  Prosody Encoded in the Neural Audio Codec Mimi?}
\name{Nicolas Ballier$^1$, Artem Saloev $^1$,  Erin Pacquetet$^2$, Taylor Arnold$^3$}

\address{$^1$ ALTAE, Université Paris Cité, F-75013 Paris, France\\
  $^2$ SCIAM, 10 rue de Penthièvre, F-75008 Paris, France\\
  $^3$ Linguistics \& Cognitive Science, University of Richmond, USA}

\email{nicolas.ballier@u-paris.fr, artem.saloev@gmail.com, erin.pacquetet@sciam.fr, tarnold2@richmond.edu}

\begin{document}

\maketitle
\begin{abstract}
  Neural audio codecs (NACs) have become a fundamental infrastructure for modern speech-language models, discretizing continuous speech signals into token sequences that enable large language models to process and generate audio. Most research evaluates codec quality at the decoding phase, assessing naturalness of reconstructed speech. However, little research investigates what phonetic information is captured during encoding. In this paper, we investigated the potential correlates of prosody in the NAC Mimi, a codec used by conversational speech agents such as Moshi and Sesame and used in downstream tasks such as speech-to-speech translation (Hibiki). We analyzed the eight codebooks generated by Mimi when encoding incrementally resynthesized English speech (pitch, duration, and glissando value of the last syllable). We investigated the potential prosodic correlates of the codebooks using Canonical Correlation Analysis (CCA) of the different codebooks generated by Mimi, and assessed the correlation between the codebook values and the resynthesized version. We show the overimportance of the first codebook (codebook\_0) compared to the others and that it clearly correlates with segmental features; the CCA shows moderate correlation between the prosodic features and the other codebooks used to represent speech.  
\end{abstract}

\section{Introduction}

Neural audio codecs (NACs) have become central in the design of modern speech models, enabling efficient audio tokenization. These models use neural networks to compress speech into discrete code sequences for high-quality reconstruction at low bitrates \cite{kyutai2024moshi}. Many codecs have been developed for industry needs, especially in conversational agents and specific applications. For instance, NotebookLM is an application from Google that leverages the Soundstream codec to produce podcasts \cite{zeghidour2021soundstream}. We can also cite Google Research’s AudioLM model \cite{borsos2023audiolm}, an audio generation model, that employs SoundStream’s discrete representations as targets during its training process. In comparison, Meta Research’s VALL‑E utilizes the Encodec codec \cite{defossez2023high} as an intermediate feature for text‑to‑speech tasks. Neural models such as Vall-E \cite{wang2023neural} have been used as language modeling tools for tasks like text-to-speech (TTS) synthesis and speech-to-speech neural translation. Vall-E's architecture demonstrates the potential of neural vocoders for accurate voice synthesis by efficiently encoding speaker identity, prosody, and linguistic content into compressed representations. Finally, \cite{zhang2023speak} introduced VALL-E X, a cross-lingual neural codec language model specifically designed for multilingual speech synthesis. VALL-E X leverages large-scale training data to facilitate cross-lingual transfer, enabling high-quality speech generation across languages with minimal speaker-specific data.

In this paper, we focus on Mimi, a codec designed for the conversational agent Moshi \cite{kyutai2024moshi}, also used by the conversational agent Sesame AI \footnote{\url{https://www.ai-sesame.com/}} and the speech-to-speech translation system Hibiki \cite{labiausse2025highfidelity}. The NAC Mimi uses Residual Vector Quantization (RVQ) to generate hierarchical discrete speech representations \cite{zhao2025prosody}. RVQ not only compresses phonetic information, but may also distill semantic and suprasegmental speech features. In the encoding phase, Mimi is reported to encode the semantic representation of speech in the first codebook, using distillation from WaveLM. The other codebooks are specialized for ``acoustic representations'' \cite{kyutai2024moshi}. These range from codebook\_0 to codebook\_7, in a process referred to as semantic ``distillation". We aim to discuss this division of labour between the first codebook specialized in segmental representation and the seven other codebooks, apparently the most likely to capture prosodic variability. Our paper aims to verify this, testing the correlations between the 2048 tokens used in the 8 codebooks \cite{kyutai2024moshi} with phone posteriorgram predictions and three prosodic features that we have systematically modified using the WinPitch software \cite{martin2012multi}. The features we chose to modify incrementally are duration, pitch, and glissando values. Duration and pitch are core dimensions of prosodic structure, while the glissando threshold was used as a way to triangulate pitch value and duration. 

Most codec studies focus on decoding quality, with limited research on encoding behavior or codebook properties. Little research has been carried out on the encoding side, especially when trying to find correlates in the codec codebooks when investigating the properties of the speech tokenizer codebook. Others \cite{cao2020code}, \cite{Churchwell2024} use the posteriorgram method to try to find correspondence between the probability of detecting a phone and the codebook. To the best of our knowledge, no systematic investigation has been carried out in the prosodic domain.

The rest of the paper is organized as follows: Section \ref{sec:prev_res} presents the main studies on Neural Audio Codecs. Section \ref{sec:methods} explains how we have analyzed the Mimi codebooks. Section \ref{sec:results} presents our results, Section \ref{sec:discussion} discusses them and Section \ref{sec:conclusion}  suggests future research and concludes.

\section{Previous Research}
\label{sec:prev_res}

\subsection{The multiplication of codecs}
Research of phonetic features in codecs is still an emergent field compared to a more established literature on the probing of phonetic representation in speech models like Wav2vec  \cite{english2022domain} or Whisper \cite{ballier2024probing,liang2025beyond}. 
\cite{onda2025benchmarking} analyzed tokens from four self-supervised learning models (HuBERT, ContentVec, data2vec, emotion2vec) manipulating prosody and speaker identity from the TIMIT corpus \cite{garofolo1993timit} to observe their sensitivity to prosody.
Recently, \cite{ji2025language} focused on discrete acoustic codec space analysis, still emphasizing speech reconstruction quality. They analyze audio restoration in clean and noisy conditions and compare various codecs. They present their own codec (Language-codec), devised using a Masked Channel Residual Vector Quantization (MCRVQ) mechanism to limit what is learnt on the first three channels (codebooks)\footnote{In the literature that "codebook" refers to the successions of tokens produced when encoding the sound file, and to the inventory of the tokens (in our case, from 0 to 2,047) assigned through the RQV process to a corresponding frame of the acoustic space.} and adds a fourth codebook using Short-Time Fourier Transform.
 
 \cite{sadok25_interspeech} employed a masked encoder \cite{sadok2025ancogen} to investigate pitch representations in NACs. They visualized encoder internal representations with t-SNE, examining pitch feature encoding and layer organization. For voice conversion and prosody manipulation, \cite{zhao25d_interspeech} developed a prosody-aware audio codec encoder (PACE). The model explicitly conditions encoding on prosodic features, enabling fine control in speech synthesis or modification.
\subsection{Probing codec representations}
Most neural audio codec analyses focus on decoding \cite{muller24c_interspeech}, \cite{mack2025assessing}, using mean opinion score (MOS) to mimic human speech quality judgments. MOS is the standard benchmark for evaluating codec speech reconstruction quality. Human ratings typically range from 1 (bad) to 5 (excellent) and assess reconstructed .wav quality, while automatic speech prediction systems \cite{saeki2022utmos} emerge for codec benchmarking. Another decoding quality metric is word error rate, measured from reconstructed signals used in speech recognition \cite{zhang2023speechtokenizer}.

In contrast, encoding-focused studies often align codec codebooks to phoneme-like units via ABX methodology \cite{schatz13_interspeech}, testing minimal pair discrimination. Another approach uses segmental realization with datasets like TIMIT to study encoder and speech decoder performance. For example, posteriogram techniques  \cite{churchwell2024high} estimate what was in the signal and how it could correspond to individual tokens of the codebook. This enables calculation of metrics emphasizing the frequency of codebook token usage matched to segmental units. However, systematic analyses relating to acoustic tokens and their corresponding values remain limited. To address this, benchmarks such as AnCoGen  \cite{sadok2025ancogen} have been proposed to provide comprehensive evaluations, incorporating supra-segmental features alongside segmental ones. These frameworks capture and analyze prosodic correlates within encoded speech representations.

A downside is that acoustic property studies often coincide with new codecs (e.g., FlexiCodec \cite{li2025flexicodec}, Language-Codec \cite{ji2025language}), making it harder to maintain a consistent analysis. For example, \cite{yang2023hifi} proposed an audio codec toolkit to facilitate research but also produced their HiFi-Codec. Research on phonetic correlates has mostly focused on segmental features to the detriment of prosodic features, and, to the best of our knowledge, no canonical correlation analysis has been proposed for codecs to investigate prosodic correlates.

\subsection{Residual Vector Quantization and Semantic Distillation}
Most recent codecs resort to Residual Vector Quantization (RVQ), which is a hierarchical compression technique used by neural audio codecs such as Mimi. RVQ quantizes residual speech signals across multiple sequential layers of codebooks, each layer encoding information not captured by the preceding layers \cite{chen2022wavlm}. This layered architecture produces a discrete representation that includes both segmental phonetic details and higher-level acoustic and semantic features. Semantic distillation denotes the hypothesis that the composite representation derived from RVQ codebooks implicitly encodes semantic and suprasegmental speech aspects.

\section{Material and methods}
\label{sec:methods}

In our experiments, we aim to identify which aspects of the speech signal are encoded by the Mimi neural audio codec's codebooks. The main research question we aim to answer is: are the speech correlates more segmental or suprasegmental? 

\subsection{Methods}
 Mimi is described as producing one ``semantic" codebook (codebook\_0), while the remainder of codebooks are "acoustic" tokens. Our study evaluates this hypothesis with two experiments, first by correlating codebook embeddings with prosodic features (pitch, duration, glissando) and second with posteriorgram phone predictions. We performed a Canonical Correlation Analysis (CCA) for the prosodic features as well as a classification task to see if we can predict the correlates of speech with the codebook token values.

\subsubsection{Canonical Correlation Analysis (CCA)}
Canonical Correlation Analysis (CCA) was employed to investigate the relationship between the multidimensional neural codec codebooks and prosodic features derived from resynthesized speech. The goal is to evaluate if prosodic information is encoded in the neural codec's discrete representations. CCA identifies linear combinations of variables from two datasets that are maximally correlated, allowing exploration of shared representations between acoustic features and codec embeddings \cite{hardoon2004canonical}. Because the codebook tokens were categorical (with token IDs ranging from 1 to 2048), we resorted to kernel CCA using the R \cite{RCoreTeam2024} package kernel \cite{Karatzoglou2004kernlab}. We defined a Hamming kernel (which counts the proportion of matches across the codebook vectors) for the codebook categorical data, and measured the correlation with the three quantitative prosodic data (pitch, duration, glissando values).

\subsubsection{Testing Mimi Representations}
Previous work \cite{kyutai2024moshi}, investigating the Mimi neural audio codec’s encoder, have primarily employed the XAB perceptual method to analyze its phonetic correlates. We have reasons to believe that this method fails to capture the semantic codebook. Mimi produces a token every 80 milliseconds, which may be about the segment or below the segment for long vowels. We would rather hypothesize that the different token IDs for the semantic codebook representation are more likely to correspond to a hybrid combination of monophones, diphones, triphones, and even quadphones. Our motivation for including these phonetic and suprasegmental correlates in our analysis arises from the hypothesis that Mimi's codebook tokens may also encode connected speech processes. These may include phenomena such as assimilations or persistent repetitive segments, which are characteristic of natural fluent speech. 

We analyze the capability of codebooks to capture segmental representations by comparing phone predictions derived from posteriorgrams \cite{churchwell2024high} with the discrete tokens generated by the Mimi codec, and we try to investigate whether the sequence of the eight codebooks can be correlated to the phone prediction obtained by the posteriorgrams (which are the generation of posterior probabilities of the phonetic categories). We aligned the posteriorgram predictions (the phone predicted with the highest probability, one every 20ms) to each time frame corresponding to a Mimi codebook (80ms) for the 6300 files of the TIMIT data. We excluded the NA and [PAD] values, resulting in a dataset of  52,749 Mimi tokens with their four corresponding posteriorgram predictions. The Mimi codec allows for 8, or 32 codebooks. We focused on the 8 version but also computed results for the 32 codebook version for our systematic investigation of the prosodic features and of the segmental features. 

\subsubsection{Classification Task}
We used the Tilburg memory-based learner (TiMBL) that implements the k-nn algorithm for categorical values as predictors \cite{daelemans2004timbl} to analyze whether we could predict our three prosodic features or the posteriorgram phone labels with the codebook token values. TiMBL operates by storing labeled instances and classifying new data points based on the majority class among their k closest neighbors in the feature space, where closeness is typically measured by a distance metric. In our study, TiMBL is applied to classify tokens generated by the Mimi codec to assess how well discrete codebook embeddings can represent phonetic and prosodic classes. To be more specific, we only analyzed the tokens of the codebooks that corresponded to the synthetically modified syllable. In 75\% of the cases (the longest files), the last five tokens corresponded to the final silence and the eight previous tokens matched the representation of the modified syllables. For the prosodic predictions, we used TiMBL with 64 features, corresponding to eight tokens multiplied by eight codebooks.   

\subsection{Data}
For the investigation of segmental correlates, we replicate the codec analysis proposed by the analysis of the SpeechTokenizer codec \cite{zhang2023speechtokenizer} using the TIMIT dataset. The TIMIT dataset consists of broadband recordings from 630 speakers of eight major American English dialects, each reading ten phonetically rich sentences with time-aligned transcriptions\cite{garofolo1993timit}. We probed the posteriorgram correspondence between the phone predicted by posteriorgram generation with a system trained on the TIMIT datasets  \cite{churchwell2024high} and the Mimi codebooks that were generated using the TIMIT data.

For the investigation of prosodic correlates, we utilized a synthetic dataset created using the WinPitch software \cite{martin2012multi}. We controlled for pitch, duration, and glissando for the modification of the last syllable of the sentence ``He arrived on time". We systematically modified the pitch (+10Hz), duration (+10ms), and glissando values in regular increments, allowing for the following variations to be observed. We generated the sentences with a value of pitch ranging from 100 to 490 Hz and the duration of the last syllable range from 10 to 390 milliseconds. 1029 variants of prosodic events in total were analyzed and compared to the eight vectors of the corresponding Mimi codebooks.

\subsubsection{Artificial generation of speech with codebook token IDs}
The Mimi codebook generates sequences of tokens on eight codebooks (or more), which are the representations of the  speech signal. We also tested the potential correlates of the Mimi token codebooks by artificially generating codebook vectors with arbitrary token IDs values.

\section{Results}
\label{sec:results}
\subsection{Mixed results}
For our CCA analysis, about 25\% of shared variance ($\approx$ 0.498² = 0.248) is captured by the first canonical dimension. We did not resort to Wilks test as it would be just an approximation since we are using a kernel method (kernel-based similarity, not raw numeric data). We then ran a classification task to realize that the eight codebooks were rather poor predictors of pitch (and subsequently, glissando) (see Table~\ref{tab:timbl-codebook-comparison}) but better predictors of the phone labels predicted by the posteriorgram method. For the phone predictions, the tokens have been realigned to the corresponding time stamps, so that eight tokens from eight codebooks are used as predictors.

\begin{table}[ht]
\centering
\begin{tabular}{lcc}
\hline
\textbf{Metric}            & \textbf{8 codebooks} \\
\hline
Duration accuracy          & 92.3\%  \\
Glissando  accuracy                 & 8.1\%  \\
Pitch accuracy                     & 15.5\%  \\
Phone accuracy             & 68.32\%  \\
\hline
\end{tabular}
\caption{TiMBL prediction performance (with K=5) for the 8 tokens of the 8 codebooks for our three prosodic variables and phone predictions with the aligned token for the TIMIT data} 
\label{tab:timbl-codebook-comparison}
\end{table}

We evaluated how Mimi encodes three prosodic parameters manipulated in the synthesized corpus: duration, pitch, and glissando. For each parameter, TiMBL was used with Leave-One-Out testing, $k=5$, Overlap metric and GainRatio weighting.

\subsubsection{Prosodic variables}
Duration classification reached 92.3\% accuracy.  
This indicates that Mimi’s tokenization preserves duration noticeably well.  
GainRatio showed that the most predictive features were concentrated in the codebooks\_1-3.  
Pitch classification yielded 15.5\% accuracy.  
Although much lower than the duration, the performance is above chance (given the number of classes to predict), demonstrating that pitch information is partially recoverable from the tokens.  
The highest GainRatio values again came from codebooks\_0–3. 

Glissando classification reached 8.1\% accuracy. Despite the fine-grained target space, glissando direction and slope were recoverable above chance. As with pitch, the strongest predictors were found in codebooks\_1-3. The last two tokens in each codebook were consistently the most predictive across all prosodic variables. 

\subsection{Number of Codebooks}
We tested two types of implementations, the 8-codebook and the 32-codebook, the hypothesis being that with the 8-codebook version we would have a better way to predict the correlates of the system and that the 8-codebook version was more likely to predict the prosodic features that we tested. The first 8 codebooks were identical in the 8 and the 32 version.

\subsection{Semantic Leakage}
Using the information gain ratio as a criterion in the classification task using TiMBL, we observed that the feature importance of the different codebooks varied systematically for all the predictions. An important observation, which perhaps mirrors RQV is the number of unique tokens grows with each new codebook. The number of unique tokens in codebook\_0 is potentially consistent with allophone representations of the syllable "time", but tokens in the codebook\_0 fail to predict the phone labels as accurately as the 8 codebooks taken together, so that some form of the semantic leakage might be at work over the other seven codebooks (see Table~\ref{tokens}). 

\begin{table}[ht]
\centering
\begin{tabular}{lr}
  \hline
\textbf{Codebook ID} & \textbf{Count} \\ 
  \hline
codebook\_0 & 27 \\ 
  codebook\_1 & 163 \\ 
  codebook\_2 & 409 \\ 
  codebook\_3 & 461 \\ 
  codebook\_4 & 577 \\ 
  codebook\_5 & 649 \\ 
  codebook\_6 & 749 \\ 
  codebook\_7 & 823 \\ 
   \hline
\end{tabular}
\caption{Number of unique token IDs per codebook} 
\label{tokens}
\end{table}

\subsection{Segmental Correlates}
Partially replicating the SpeechTokenizer experiments matching codebook IDs with posteriorgrams, we managed to match codebook IDs to posteriogram predictions. We surmise that the representation in tokens, which spans over 80 milliseconds may correspond to quadphones, triphones, diphones, and single phone representations for the 2048 token IDs of codebook\_0. 
On the TIMIT dataset, for the 80 ms time frames corresponding to the 52,749 tokens produced in codebook\_0, we observed 33,540 realisations of identical posteriogram predictions (possibly phones), 18,795 realisations of two different posteriogram predictions (possibly diphones), 413 realisations of three different posteriogram predictions (possibly triphones) and one candidate for quadphone (the plausible frequent phone sequence \textipa{/InD@/}, corresponding to tokenID  1047).

\section{Discussion}
\label{sec:discussion}

\subsection{Gender bias or specialization?} 

\begin{figure}[t]
    \centering
    \includegraphics[width=0.73\linewidth]{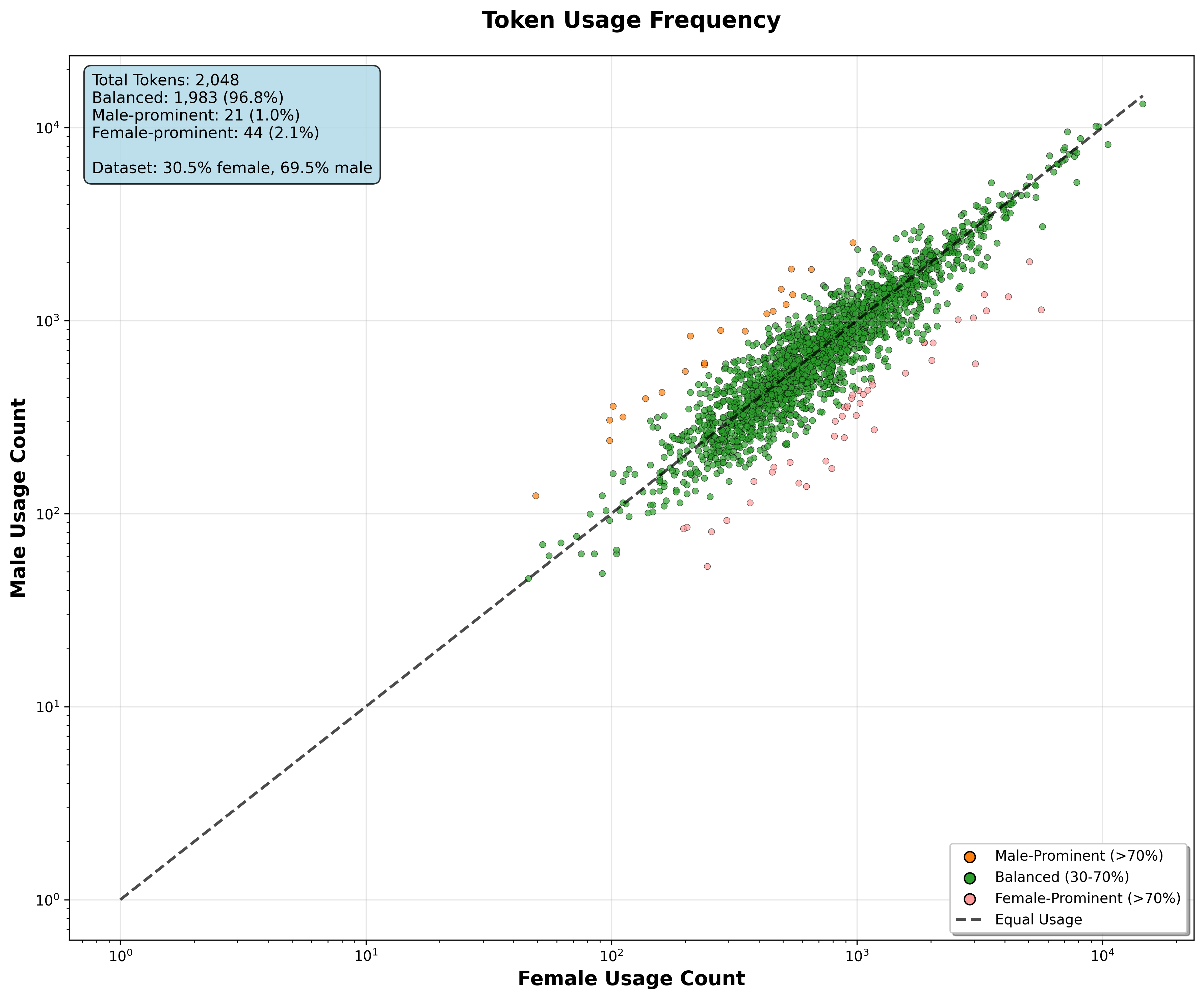}
    \caption{Gender prominence in the Mimi tokens (TIMIT data)}
    \label{fig:gender-bias}
\end{figure}

\begin{table}[t]
\centering
\begin{tabular}{cccc}
\toprule
\textbf{Codebook} & \textbf{Total Tokens} & \textbf{Exclusive Tokens} & \textbf{ratio} \\
\midrule
1 & 858 & 265 & 30.90\% \\
2 & 1,641 & 362 & 22.10\% \\
0 & 1,821 & 165 & 9.10\% \\
3 & 1,857 & 176 & 9.50\% \\
4 & 1,919 & 142 & 7.40\% \\
5 & 1,916 & 128 & 6.70\% \\
6 & 1,933 & 145 & 7.50\% \\
7 & 1,943 & 111 & 5.70\% \\
\bottomrule
\end{tabular}
\caption{Statistics of codebook tokens and exclusivity rate.}
\label{tab:codebook_stats}
\end{table}

The 2,048 tokens used in each codebook may represent distinct acoustic features for each codebook. When looking at codebook specific tokens, the one with the most exclusivity  per gender is by far codebook\_1. For instance, no female has token\_ID  6 in codebook\_1, which makes it gender exclusive for this codebook, but we see  token\_ID 6 in female data elsewhere. It is also interesting to note that codebook\_1 has far fewer unique  tokens than other codebooks. When looking at unique token (independent from the codebook they are from), a few tokens show a gender bias (none are exclusive, though).

\subsection{Ablation experiments and synthetic codebook values}

To test the meaning of the first codebook, we performed a perceptual ablation experiment replacing the first codebook token values by arbitrary numbers between 1 and 100 and the sound file was audibly very similar to the original sound file, so that we may conclude that the first codebook is surprisingly not as important as the combination of all the codebooks taken together.
When multiple 8 codebooks were simultaneously modified with synthetic or cycled values, the speech degradation was noticeable, characterized by noise and unintelligible words. This suggests that the speech decoder relies heavily on joint coherence across multiple codebooks to maintain naturalness and clarity.

\subsection{Specialization or Distributed semantics?}
An alternative experiment is the resynthesis of the codebooks with the assumption that if the codebooks have specific meanings or prosodic values associated to the tokens, we would be expected to observe variability in frequency, pitch or glissando investigating the output from codebooks synthetically generated using artificial numbers from zero to 2,048. The sequences are somehow arbitrary. We suspect that there is a form of solidarity for the eight values of the codebook, so that we also tested on a minimal number of token units that we found in the data the change that we observed modifying these token ideas.

\section{Conclusion}
\label{sec:conclusion}
In this paper, we have shown that the division of labour for the Mimi codec between the semantic codebook and the acoustic codebooks might not be as clear-cut as initially posited \cite{kyutai2024moshi}. The so-called `acoustic codebooks' (from codebook\_1 to codebook\_7) correlate strongly with duration, but not with glissando and pitch values. It might be the case that tokens are good predictors of duration because, mechanically, as duration increases, tokens are produced every 80 ms. However, clearer correlations can be established with the codebook\_0 values with phone, diphone, triphone and quadphone representations than suggested in the ABX discrimination method \cite{schatz2016abx,kyutai2024moshi}. Future neural audio codecs may benefit from a more specialized allocation of representational capacity, where different vectors are optimized for distinct prosodic dimensions. This could pave the way for better prosodic controllability in synthesis and robustness in speech models.


\bibliographystyle{IEEEtran}
\bibliography{mybib}

@article{hardoon2004canonical,
  title={Canonical correlation analysis: An overview with application to learning methods},
  author={Hardoon, David R and Szedmak, Sandor and Shawe-Taylor, John},
  journal={Neural computation},
  volume={16},
  number={12},
  pages={2639--2664},
  year={2004},
  publisher={MIT Press}
}

@techreport{daelemans2004timbl,
  title = {{TiMBL}: Tilburg Memory Based Learner, version 6.0, Reference Guide},
  author = {Daelemans, Walter and Zavrel, Jakub and van der Sloot, Ko and van den Bosch, Antal},
  institution = {ILK Research Group, Tilburg University},
  year = {2004},
  address = {Tilburg, The Netherlands},
  url = {https://ilk.uvt.nl/timbl/}
}

@techreport{kyutai2024moshi,
    author = {Alexandre D\'efossez and Laurent Mazar\'e and Manu Orsini and Am\'elie Royer and Patrick P\'erez and Herv\'e J\'egou and Edouard Grave and Neil Zeghidour},
    title = {Moshi: a speech-text foundation model for real-time dialogue},
    institution = {Kyutai},
    year={2024},
    month={September},
    url={http://kyutai.org/Moshi.pdf},
}

@article{schatz13_interspeech,
  title     = {Evaluating speech features with the minimal-pair {ABX} task: analysis of the classical MFC/PLP pipeline},
  author    = {Thomas Schatz and Vijayaditya Peddinti and Francis Bach and Aren Jansen and Hynek Hermansky and Emmanuel Dupoux},
  year      = {2013},
  booktitle = {Interspeech 2013},
  pages     = {1781--1785},
  doi       = {10.21437/Interspeech.2013-441},
  issn      = {2958-1796},
}

@article{zeghidour2021soundstream,
  title={Soundstream: An end-to-end neural audio codec},
  author={Zeghidour, Neil and Luebs, Alejandro and Omran, Ahmed and Skoglund, Jan and Tagliasacchi, Marco},
  journal={IEEE/ACM Transactions on Audio, Speech, and Language Processing},
  volume={30},
  pages={495--507},
  year={2021},
  publisher={IEEE}
}

@article{saeki2022utmos,
  title={{UTMOS: UTokyo-SaruLab System for VoiceMOS Challenge 2022}},
  author={Saeki, Takaaki and Xin, Detai and Nakata, Wataru and Koriyama, Tomoki and Takamichi, Shinnosuke and Saruwatari, Hiroshi},
  booktitle={Proceedings of the Annual Conference of the International Speech Communication Association, INTERSPEECH},
  volume={2022},
  pages={4521--4525},
  year={2022}
}

@article{sadok2025ancogen,
  title={AnCoGen: Analysis, control and generation of speech with a masked autoencoder},
  author={Sadok, Samir and Leglaive, Simon and Girin, Laurent and Richard, Ga{\"e}l and Alameda-Pineda, Xavier},
  journal={ICASSP 2025-2025 IEEE International Conference on Acoustics, Speech and Signal Processing (ICASSP)},
  pages={1--5},
  year={2025},
  organization={IEEE}
}

@article{garofolo1993timit,
  title={TIMIT acoustic-phonetic continuous speech corpus},
  author={Garofolo, John S and Lamel, Lori F and Fisher, William M and Pallett, David S and Dahlgren, Nancy L and Zue, Victor and Fiscus, Jonathan G},
  journal={NASA STI/Recon technical report n},
  year={1993},
  publisher={Linguistic data consortium}
}

@article{chen2022wavlm,
  title={WavLM: Large-Scale Self-Supervised Pre-Training for Full Stack Speech Processing},
  author={Chen, Sanyuan and Wang, Chengyi and Chen, Zhengyang and Wu, Yu and Liu, Shujie and Chen, Zhuo and Li, Jinyu and Kanda, Naoyuki and Yoshioka, Takuya and Xiao, Xiong and others},
  journal={IEEE Journal of Selected Topics in Signal Processing},
  volume={16},
  number={6},
  pages={1505},
  year={2022}
}

@article{zhang2023speechtokenizer,
  title={SpeechTokenizer: Unified Speech Tokenizer for Speech Large Language Models},
  author={Zhang, Xin and Zhang, Dong and Li, Shimin and Zhou, Yaqian and Qiu, Xipeng},
  journal={CoRR},
  year={2023}
}

@article{defossez2023high,
  title={High Fidelity Neural Audio Compression},
  author={D{\'e}fossez, Alexandre and Copet, Jade and Synnaeve, Gabriel and Adi, Yossi},
  journal={Trans. Mach. Learn. Res.},
  year={2023}
}

@inproceedings{sadok25_interspeech,
  title     = {{Bringing Interpretability to Neural Audio Codecs}},
  author    = {Samir Sadok and Julien Hauret and Éric Bavu},
  year      = {2025},
  booktitle = {{Interspeech 2025}},
  pages     = {5023--5027},
  doi       = {10.21437/Interspeech.2025-115},
  issn      = {2958-1796},
}

@article{zhao2025prosody,
  title={Prosody-Adaptable Audio Codecs for Zero-Shot Voice Conversion via In-Context Learning},
  author={Zhao, Junchuan and Wang, Xintong and Wang, Ye},
  journal={arXiv preprint arXiv:2505.15402},
  year={2025}
}

@article{wang2023neural,
  title={Neural codec language models are zero-shot text to speech synthesizers},
  author={Wang, Chengyi and Chen, Sanyuan and Wu, Yu and Zhang, Ziqiang and Zhou, Long and Liu, Shujie and Chen, Zhuo and Liu, Yanqing and Wang, Huaming and Li, Jinyu and others},
  journal={arXiv preprint arXiv:2301.02111},
  year={2023}
}

@article{zhang2023speak,
  title={Speak foreign languages with your own voice: Cross-lingual neural codec language modeling},
  author={Zhang, Ziqiang and Zhou, Long and Wang, Chengyi and Chen, Sanyuan and Wu, Yu and Liu, Shujie and Chen, Zhuo and Liu, Yanqing and Wang, Huaming and Li, Jinyu and others},
  journal={arXiv preprint arXiv:2303.03926},
  year={2023}
}

@inproceedings{zhao25d_interspeech,
  title     = {{Prosody-Adaptable Audio Codecs for Zero-Shot Voice Conversion via In-Context Learning}},
  author    = {Junchuan Zhao and Xintong Wang and Ye Wang},
  year      = {2025},
  booktitle = {{Interspeech 2025}},
  pages     = {4893--4897},
  doi       = {10.21437/Interspeech.2025-464},
  issn      = {2958-1796},
}

@phdthesis{schatz2016abx,
  title={{ABX}-discriminability measures and applications},
  author={Schatz, Thomas},
  year={2016},
  school={Universit{\'e} Paris 6 (UPMC)}
}

@article{yang2023hifi,
  title={HiFi-Codec: Group-residual Vector quantization for High Fidelity Audio Codec},
  author={Yang, Dongchao and Liu, Songxiang and Huang, Rongjie and Tian, Jinchuan and Weng, Chao and Zou, Yuexian},
  journal={CoRR},
  year={2023}
}

@article{borsos2023audiolm,
  title={AudioLM: A Language Modeling Approach to Audio Generation},
  author={Borsos, Zal{\'a}n and Marinier, Rapha{\"e}l and Vincent, Damien and Kharitonov, Eugene and Pietquin, Olivier and Sharifi, Matt and Roblek, Dominik and Teboul, Olivier and Grangier, David and Tagliasacchi, Marco and others},
  journal={IEEE/ACM Transactions on Audio, Speech, and Language Processing},
  volume={31},
  pages={2523--2533},
  year={2023},
  publisher={IEEE}
}

@inproceedings{ballier2024probing,
  title={{Probing Whisper predictions for French, English and Persian transcriptions}},
  author={Ballier, Nicolas and Burin, L{\'e}a and Namdarzadeh, Behnoosh and Ng, Sara B and Wright, Richard and Yun{\`e}s, Jean-Baptiste},
  booktitle={Proceedings of the 7th International Conference on Natural Language and Speech Processing (ICNLSP 2024)},
  pages={129--138},
  year={2024}
}

@inproceedings{liang2025beyond,
  title={Beyond WER: Probing Whisper’s Sub-token Decoder Across Diverse Language Resource Levels},
  author={Liang, Siyu and Ballier, Nicolas and Levow, Gina-Anne and Wright, Richard},
  booktitle={Proceedings of the 2025 Conference on Empirical Methods in Natural Language Processing},
  pages={31225--31235},
  year={2025}
}

@inproceedings{english2022domain,
  title={Domain-informed probing of wav2vec 2.0 embeddings for phonetic features},
  author={English, Patrick Cormac and Kelleher, John and Carson-Berndsen, Julie},
  booktitle={Proceedings of the 19th SIGMORPHON Workshop on Computational Research in Phonetics, Phonology, and Morphology},
  pages={83--91},
  year={2022}
}

@inproceedings{
labiausse2025highfidelity,
title={High-Fidelity Simultaneous Speech-To-Speech Translation},
author={Tom Labiausse and Laurent Mazar{\'e} and Edouard Grave and Alexandre D{\'e}fossez and Neil Zeghidour},
booktitle={Forty-second International Conference on Machine Learning},
year={2025},
url={https://openreview.net/forum?id=fgjN8B6xVX}
}

@inproceedings{martin2012multi,
  title={Multi methods pitch tracking},
  author={Martin, Philippe},
  booktitle={Proc. SpeechProsody 2012},
  pages={47--50},
  year={2012}
}

@inproceedings{ji2025language,
  title={Language-codec: bridging discrete codec representations and speech language models},
  author={Ji, Shengpeng and Fang, Minghui and Zuo, Jialong and Jiang, Ziyue and Wang, Dingdong and Wang, Hanting and Huang, Hai and Zhao, Zhou},
  booktitle={Proceedings of the 63rd Annual Meeting of the Association for Computational Linguistics (Volume 1: Long Papers)},
  pages={13332--13345},
  year={2025}
}

@article{li2025flexicodec,
  title={FlexiCodec: A Dynamic Neural Audio Codec for Low Frame Rates},
  author={Li, Jiaqi and Qian, Yao and Hu, Yuxuan and Zhang, Leying and Wang, Xiaofei and Lu, Heng and Thakker, Manthan and Li, Jinyu and Zhao, Shang and Wu, Zhizheng},
  journal={arXiv preprint arXiv:2510.00981},
  year={2025}
}

@article{onda2025benchmarking,
  title={Benchmarking Prosody Encoding in Discrete Speech Tokens},
  author={Onda, Kentaro and Fukayama, Satoru and Saito, Daisuke and Minematsu, Nobuaki},
  journal={arXiv preprint arXiv:2508.11224},
  year={2025}
}

@INPROCEEDINGS{Churchwell2024,
  author={Churchwell, Cameron and Morrison, Max and Pardo, Bryan},
  booktitle={2024 IEEE International Conference on Acoustics, Speech, and Signal Processing Workshops (ICASSPW)}, 
  title={High-Fidelity Neural Phonetic Posteriorgrams}, 
  year={2024},
  volume={},
  number={},
  pages={823-827},
  doi={10.1109/ICASSPW62465.2024.10669905}}

@inproceedings{cao2020code,
  title={Code-switched speech synthesis using bilingual phonetic posteriorgram with only monolingual corpora},
  author={Cao, Yuewen and Liu, Songxiang and Wu, Xixin and Kang, Shiyin and Liu, Peng and Wu, Zhiyong and Liu, Xunying and Su, Dan and Yu, Dong and Meng, Helen},
  booktitle={ICASSP 2020-2020 IEEE International Conference on Acoustics, Speech and Signal Processing (ICASSP)},
  pages={7619--7623},
  year={2020},
  organization={IEEE}
}

@article{mack2025assessing,
  title={Assessing speech quality metrics for evaluation of neural audio codecs under clean speech conditions},
  author={Mack, Wolfgang and Topaloglu, Nezih and Lechler, Laura and Bali{\'c}, Ivana and Craciun, Alexandra and Yesilbursa, Mansur and Wojcicki, Kamil},
  journal={arXiv preprint arXiv:2509.24457},
  year={2025}
}

@inproceedings{muller24c_interspeech,
  title     = {{Speech quality evaluation of neural audio codecs}},
  author    = {Thomas Muller and Stephane Ragot and Laetitia Gros and Pierrick Philippe and Pascal Scalart},
  year      = {2024},
  booktitle = {{Interspeech 2024}},
  pages     = {1760--1764},
  doi       = {10.21437/Interspeech.2024-1072},
  issn      = {2958-1796},
}

@inproceedings{churchwell2024high,
  title={High-fidelity neural phonetic posteriorgrams},
  author={Churchwell, Cameron and Morrison, Max and Pardo, Bryan},
  booktitle={2024 IEEE International Conference on Acoustics, Speech, and Signal Processing Workshops (ICASSPW)},
  pages={823--827},
  year={2024},
  organization={IEEE}
}

@Article{Karatzoglou2004kernlab,
    title = {kernlab -- An {S4} Package for Kernel Methods in {R}},
    author = {Alexandros Karatzoglou and Alex Smola and Kurt Hornik and Achim Zeileis},
    journal = {Journal of Statistical Software},
    year = {2004},
    volume = {11},
    number = {9},
    pages = {1--20},
    doi = {10.18637/jss.v011.i09},
  }

@Manual{RCoreTeam2024,
    title = {R: A Language and Environment for Statistical Computing},
    author = {{R Core Team}},
    organization = {R Foundation for Statistical Computing},
    address = {Vienna, Austria},
    year = {2024},
    url = {https://www.R-project.org/},
  }

\end{document}